\documentclass[letterpaper, 10 pt, conference]{ieeeconf}

\IEEEoverridecommandlockouts
\usepackage{times}
\usepackage{amsmath}
\usepackage{amssymb}
\usepackage{graphicx}

\usepackage{array}
\usepackage{xspace}
\usepackage{booktabs}
\usepackage{algorithm}
\usepackage{algpseudocode}

\makeatletter
\renewcommand\fs@ruled{%
  \def\@fs@cfont{\bfseries}\let\@fs@capt\floatc@ruled
  \def\@fs@pre{\vspace*{5pt}\hrule height.8pt depth0pt \kern2pt}%
  \def\@fs@post{\kern2pt\hrule\relax}%
  \def\@fs@mid{\kern2pt\hrule\kern2pt}%
  \let\@fs@iftopcapt\iftrue}
\makeatother

\makeatletter
\def\ps@symboticcopyright{%
  \def\@oddhead{}%
  \def\@evenhead{}%
  \def\@oddfoot{%
    \hfil{\footnotesize
    \textcopyright~2026 Symbotic LLC. All rights reserved.}\hfil}%
  \let\@evenfoot\@oddfoot
}
\makeatother

\newlength{\SuccessHalf}
\newcommand{\methodname}{MAGIC\xspace}
\DeclareMathOperator*{\argmin}{arg\,min}
\newcommand{\etalcite}[1]{et~al.~\cite{#1}}
\newcommand{\problemname}{Belief-Aware MAPF\xspace}

\makeatletter
\newcommand{\algmargin}{\the\ALG@thistlm}
\makeatother
\algnewcommand{\parState}[1]{%
  \State\parbox[t]{\dimexpr\linewidth-\algmargin}{\strut #1\strut}}

\title{\LARGE \bf
Belief-Aware Multi-Agent Path Finding under Map Uncertainty
}

\author{Viraj Parimi$^{*1}$, Shao-Hung Chan$^{2}$, Han Zhang$^{2}$, Jingkai Chen$^{2}$ and Brian Williams$^{1}$
\thanks{* Corresponding at {\tt\footnotesize \{vparimi\}@mit.edu}.}
\thanks{$^{1}$ Computer Science and Artificial Intelligence Laboratory, Massachusetts Institute of Technology, Cambridge, MA 01239.}
\thanks{$^{2}$ Symbotic Inc., Wilmington, MA 01887, USA}
}

\begin{document}

\maketitle
\thispagestyle{symboticcopyright}
\pagestyle{empty}

\begin{abstract}

Multi-Agent Path Finding (MAPF) aims to find collision-free paths for multiple agents in a shared environment.
Classical MAPF assumes that all static obstacles are known in advance, but real-world environments can change unexpectedly due to fallen objects, spills, or other local disturbances.
When such changes are spatially correlated, an observation can inform traversability estimates beyond the observed location.
Prior approaches address uncertainty in traversability through contingent plans or replanning based on direct observations, but do not leverage this spatial dependence to infer the traversability of nearby unobserved locations.
As a result, they cannot use one observation to anticipate nearby unobserved obstacles that may cause costly rerouting later.
We focus on \problemname, where map discrepancies are fixed during execution but initially unknown, and observations can be informative beyond the observed location.
We propose Multi-Agent Gaussian Belief Inference for Coordination (\methodname), a framework that updates a shared belief about traversability online based on agents’ observations.
\methodname uses a Gaussian Markov Random Field and Gaussian Belief Propagation to approximately infer traversability and construct detour-aware costs for standard MAPF planners.
Our experiments on MAPF benchmarks show that \methodname reduces the executed sum of costs compared to existing approaches on $96.3\%$ of instances, across several planner families and teams of up to 800 agents, demonstrating its applicability to large-scale MAPF problems.

\end{abstract}

\section{INTRODUCTION}
\label{sec:intro}

Multi-Agent Path Finding (MAPF) is the problem of finding a set of collision-free paths, one for each agent, from its start to its goal location in a shared environment~\cite{stern2019mapf}.
It provides a common abstraction for coordinating robots in warehouses and other shared spaces~\cite{wurman2008coordinating}. 
Most MAPF planners assume a fixed and accurately known environment. 
In practice, the planner's representation may not accurately reflect the environment at execution time.
For example, during automated warehouse operations, a fallen pallet, a closed passage, or a spill may make nearby locations inaccessible and invalidate planned paths.
In this work, we consider initially unknown map discrepancies that remain fixed during execution and are discovered through local observations.
Planning must therefore account not only for interactions among agents, but also for uncertainty about which parts of the environment remain traversable. 

Recent approaches have begun to relax the assumption that the environment is completely known. 
MAPF under obstacle uncertainty (MAPF-OU) constructs contingent plans that branch on future observations~\cite{shofer2023mapfou} but becomes computationally costly as uncertainty grows. 
MAPF with imperfect maps (MAPF-IM) instead interleaves planning and execution and replans as uncertain parts of the environment are directly observed~\cite{malka2024mapfim}.
However, 
certain environmental changes can exhibit spatial structure~\cite{ocallaghan2012gpom}. 
A blocked aisle or a displaced object may affect several nearby locations, so an observation at one location can provide information about the unobserved parts of the environment.
This motivates an online MAPF framework that uses spatial dependence to infer traversability beyond directly observed locations.


\begin{figure*}[t]
    \vspace*{4pt}
    \centering
    \includegraphics[width=0.9\linewidth]{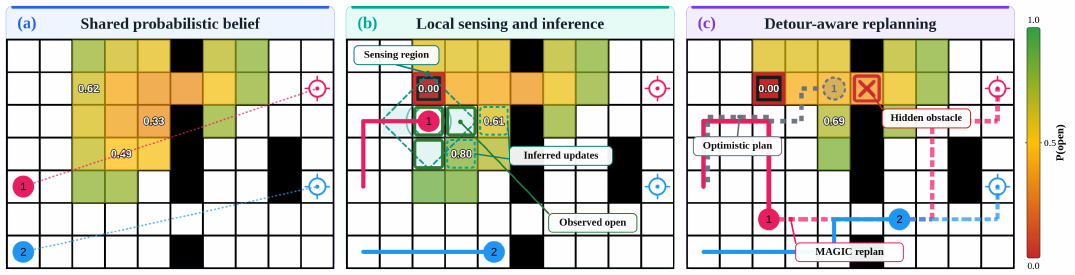}
    \caption{\textbf{Overview of \methodname.} 
    White and black locations are known to be traversable and blocked, respectively. 
    Other colors show traversability probabilities at uncertain locations. 
    Filled circles mark the positions of agents, and crosshairs mark goals, with Agent~1 in pink and Agent~2 in blue. 
    (a) Agents maintain a shared posterior over locations with uncertain traversability, with representative probabilities shown. 
    (b) Solid borders mark directly observed locations, while dashed borders mark unobserved locations whose beliefs change through GaBP.
    (c) Updated beliefs and detour distances define traversal costs for a standard MAPF planner. 
    Solid and thick dashed lines show the executed paths and planned paths, respectively.
    Under optimistic replanning, which treats uncertain locations as traversable, Agent~1 would instead be at the hollow marker, where it is about to observe the hidden obstacle marked by the red cross and replan. \methodname instead routes it through the lower corridor, avoiding that later reroute.}
    \label{fig:overall}
\end{figure*}

We formulate this setting as \problemname, where agents plan under a shared probabilistic belief over initial uncertain traversability and update that belief through agents' observations.
To address this problem, we propose \textit{Multi-Agent Gaussian belief Inference for Coordination} (\methodname), a framework that integrates spatial inference with online MAPF planning.
\methodname represents spatial correlations among uncertain locations with a Gaussian Markov Random Field (GMRF) and incorporates new observations from the agents using Gaussian Belief Propagation (GaBP)~\cite{weiss2001correctness}. 
Each observation directly resolves the states of the observed locations while also updating beliefs about nearby unobserved locations.
We combine inferred traversability with local edge-bypass distances to construct detour-aware planning costs.
This allows a planner to distinguish uncertain transitions with inexpensive alternatives from those that would require a substantial detour.
We apply these traversal costs to MAPF planners and update them online as agents gather new observations, following an interleaved planning-and-execution pipeline.
Fig.~\ref{fig:overall} shows the overall framework of MAGIC.

Our contributions are threefold:

\begin{itemize}
    \item We formulate \problemname, a one-shot MAPF problem in a fixed environment where agents lack a priori knowledge of the traversability of some locations. 
    We frame it as a planning problem under a shared probabilistic belief that captures the spatial dependence among uncertain locations and quantifies their traversability.
    \item We propose \methodname, a framework that leverages this shared probabilistic belief to construct detour-aware traversal costs, enabling MAPF planners to weigh the likelihood that a planned transition remains available against the cost of an alternative route if it does not.
    \item We evaluate \methodname across standard MAPF benchmark maps, planner families, and teams of up to 800 agents, measuring executed cost, success rate, and sensitivity to uncertainty and prior information.
    Compared with existing approaches, \methodname achieves lower executed SoC on $96.3\%$ of instances.
\end{itemize}

\section{RELATED WORK}
\label{sec:relwork}

\subsection{MAPF Under Map Uncertainty and Partial Observability}

MAPF-OU constructs contingent plans that branch on observations of initially unknown obstacles, with computational cost increasing as that uncertainty grows~\cite{shofer2023mapfou}.
MAPF-IM instead interleaves planning and execution, replanning as direct observations revise assumed traversability~\cite{malka2024mapfim}.
Neither approach, however, uses spatial dependence to infer traversability of unobserved locations.

\subsection{Adaptive Guidance and Incremental Execution}

Guidance graph methods steer multi-agent traffic through edge weights without replacing the underlying planner. 
One such method, Guidance Graph Optimization, optimizes these weights to improve lifelong MAPF throughput~\cite{zhang2024ggo}.
Such methods establish weighted graphs as a practical interface for influencing the routes produced by MAPF planners.
However, their weights shape traffic rather than encode uncertainty in environmental traversability.
Interleaved planning and execution is likewise well studied. 
Rolling-Horizon Collision Resolution repeatedly replans in lifelong settings while resolving collisions only within a bounded horizon~\cite{li2021rhcr}. 
Similarly, MAPF-IM defers distant conflicts that may become irrelevant after new observations~\cite{malka2024mapfim}. 
These approaches show that repeated replanning can avoid committing computation to portions of a plan that may change before execution. 

\subsection{Planning With Uncertain Traversability}

Single-agent planning on uncertain graphs provides an important foundation for reasoning about decisions whose outcomes depend on unknown traversability. 
The Canadian Traveller Problem seeks policies that minimize expected travel cost when edge states are revealed during execution~\cite{nikolova2008ctp}. 
Such works show how uncertainty and the cost of recovery shape route selection, but for a single agent rather than for collision-free coordination among many.

Planning under uncertain traversability has also considered information sharing and the construction of alternative routes.
Stadler~\etalcite{stadler2023collaborative} build team policies from macro-actions that combine movement, sensing, and waiting for information from other agents. 
Veys~\etalcite{veys2024sparse} generate sparse probabilistic roadmaps that retain uncertain shortcuts and recovery paths supporting low-expected-cost policies. 
These methods reason about uncertain traversability through specialized policies or graph representations, rather than cost-based guidance for existing MAPF planners.

\subsection{Belief-Space Planning and Spatial Map Inference}

Partially observable Markov Decision Processes reason jointly about actions, observations, and evolving beliefs. 
Online planners such as Partially Observable Monte Carlo Planning use sampling to limit the search over future actions and observations~\cite{silver2010pomcp}. 
Multi-robot belief-space planning considers future observations and collaboration when selecting trajectories~\cite{regev2016belief}, 
but the joint belief and action spaces grow rapidly with the number of agents, making large-team planning computationally demanding. 

Probabilistic mapping instead exploits spatial dependencies to infer occupancy at unobserved locations. 
Gaussian-Process occupancy maps use spatial correlations~\cite{ocallaghan2012gpom}, while MRFMap models dependencies among occupancy variables using a Markov Random Field and Loopy Belief Propagation~\cite{shankar2020mrfmap}.
Related single-agent work updates beliefs over spatially correlated blockage states during planning~\cite{zhou2025scos}. 
These approaches demonstrate that spatial structure can make local observations informative about unobserved parts of the environment. 
Using such beliefs in scalable MAPF planners with explicit collision resolution remains less explored.

\section{BELIEF-AWARE MAPF}
\label{sec:problem}


\subsection{Classical MAPF}

A classical MAPF instance consists of an undirected graph $\mathcal{G}=(\mathcal{V},\mathcal{E})$ representing the environment and a set of $n$ agents $\{a_1,\dots,a_n\}$~\cite{stern2019mapf}. 
Vertices represent locations, and edges represent allowed moves.
Each agent $a_i$ has a start location $s_i \in \mathcal{V}$ and a goal location $g_i \in \mathcal{V}$.
Time is discretized into timesteps.
We write $q_i^t \in \mathcal{V}$ for the location of agent $a_i$ at timestep $t$, with $q_i^0 = s_i$.
At each timestep, an agent either waits at its current location or moves to an adjacent location in $\mathcal{G}$. 
A joint plan must avoid vertex and edge conflicts~\cite{stern2019mapf}.
A vertex conflict occurs when $q_i^t = q_j^t$ for some timestep $t$ and some $i \neq j$, and an edge conflict occurs when $q_i^t = q_j^{t+1}$ and $q_i^{t+1} = q_j^t$ for some $i \neq j$.
Let $T_i$ denote the final arrival time after which agent $a_i$ remains at $g_i$. 
The sum of costs (SoC) of a joint plan is $\sum_{i = 1}^n T_i$. 

\subsection{Fixed Environment with Unknown Obstacles}

\problemname extends classical MAPF by allowing some locations to have initially unknown traversability.
Let $\mathcal{Z} \subseteq \mathcal{V}$ denote this initially uncertain set.
Each $z \in \mathcal{Z}$ carries a binary random variable $X_z$, where $X_z = 1$ indicates that $z$ is traversable and $X_z = 0$ that it is blocked.
We write $\mathbf{X} = (X_z)_{z \in \mathcal{Z}}$ for the joint hidden traversability state and $\mathbf{x} = (x_z)_{z \in \mathcal{Z}}$ for a particular realization, i.e., the true traversability state of the uncertain locations.
Every location in $\mathcal{V} \setminus \mathcal{Z}$ is known to be traversable.
For these locations, we set $X_v = 1$ and $x_v = 1$.
Obstacles already represented in the initial graph specification are excluded from $\mathcal{V}$ altogether.
We require $s_i, g_i \in \mathcal{V} \setminus \mathcal{Z}$ for all $i$, so starts and goals are never placed at uncertain locations.
The true traversable graph induced by a particular realization $\mathbf{x}$ is denoted by $\mathcal{G}^\star(\mathbf{x}) = (\mathcal{V}^\star(\mathbf{x}), \mathcal{E}^\star(\mathbf{x}))$ with

\begin{equation}
\begin{aligned}
    \mathcal{V}^\star(\mathbf{x}) &= \{v \in \mathcal{V} \mid x_v = 1 \} \\
    \mathcal{E}^\star(\mathbf{x}) &= \{ (u,v) \in \mathcal{E} \mid u, v \in \mathcal{V}^\star(\mathbf{x}) \}
\end{aligned}
\label{eq:truegraph}
\end{equation}

Let $\mathcal{P}$ denote the environment distribution over $\mathbf{X}$, with support restricted to realizations $\mathbf{x}$ for which $\mathcal{G}^\star(\mathbf{x})$ admits a collision-free solution. 
As part of the problem setup, a realization $\mathbf{x} \sim \mathcal{P}$ is sampled and remains fixed throughout execution.
Agents know $\mathcal{Z}$ and are given a prior distribution $p_0(\mathbf{X})$, but they do not know $\mathbf{x}$. 
The planner's prior $p_0$ need not match the environment distribution $\mathcal{P}$.
We defer the specification of $p_0$ to Sec.~\ref {sec:method} and the construction of $\mathcal{Z}$ to Sec.~\ref {sec:experiments}.

\subsection{Observations and Online Objective}

We assume that sensing is local and incurs no additional cost.
Before each joint move, every agent noiselessly observes the states of uncertain locations adjacent to its current position, and immediately shares these observations with the team.
Let $\mathcal{H}_t$ denote all observations collected through timestep $t$, so every agent conditions on the same history.
The shared posterior probability that a location $v \in \mathcal{V}$ is traversable is
\begin{equation}
    b_t(v) = \Pr\nolimits_{p_0}[X_v = 1 \mid \mathcal{H}_t]
    \label{eq:belief}
\end{equation}
When $p_0$ couples nearby locations, an observation generally shifts this posterior at locations that have not been observed yet.
A solution is a centralized online policy $\pi$ that maps the current configuration $(q_1^t, \dots, q_n^t)$ and the observation history $\mathcal{H}_t$ to a joint action, eventually bringing every agent to its goal while avoiding vertex and edge conflicts. 
Executed moves must lie in $\mathcal{G}^\star(\mathbf{x})$. 
Let $\mathrm{SoC}(\pi, \mathbf{x}) = \sum_{i = 1}^n T_i$ denote the executed SoC when policy $\pi$ operates on $\mathcal{G}^\star(\mathbf{x})$. 
We seek  
\begin{equation}
    \pi^\star \in \argmin_{\pi} \; \mathbb{E}_{\mathbf{X} \sim \mathcal{P}} \bigl[ \mathrm{SoC}(\pi, \mathbf{X}) \bigr].
    \label{eq:objective}
\end{equation}
Computing $\pi^\star$ requires contingent reasoning over possible unknown location states and observation histories, and $\mathcal{P}$ is unknown to the planner. Even related single-agent routing problems with uncertain edge states are computationally hard~\cite{nikolova2008ctp}, and the multi-agent setting additionally requires joint conflict resolution. 
For this paper, we restrict attention to one-shot MAPF with fixed start and goal positions and unknown location states. 

\section{MAGIC}
\label{sec:method}

Directly optimizing Eq.~\eqref{eq:objective} would require knowledge of $\mathcal{P}$ and contingent reasoning over possible hidden states and future observation histories.
\methodname instead approximates this online decision problem by repeatedly solving a deterministic MAPF instance informed by the observations collected so far. 
At each update, we first infer a shared belief over the traversability of locations that remain unobserved.
We then translate these beliefs into traversal costs that reflect the consequence of using uncertain parts of the environment, and provide the resulting weighted graph to a standard MAPF planner.
The resulting plan is executed until new observations trigger another belief and planning update.

\subsection{Shared Traversability Belief}

We model the shared belief over the hidden traversability state $\mathbf{X}$ using a GMRF over latent scores $\mathbf{f} = (f_v)_{v \in \mathcal{V}}$, where $f_v \in \mathbb{R}$. 
Conditioned on $\mathbf{f}$, we model the traversability state of each uncertain location $v \in \mathcal{Z}$ independently as $X_v \mid f_v \sim \mathrm{Bernoulli}(\Phi(f_v))$, where $\Phi$ denotes the standard normal cumulative distribution function. A larger latent score $f_v$ indicates a higher probability that location $v$ is traversable.
This latent field has density
\begin{equation}
    p(\mathbf{f}) \propto \exp\left( -\frac{\lambda_0}{2}\sum_{v \in \mathcal{V}}(f_v - \mu_0)^2 - \frac{\lambda_1}{2}\sum_{(u, v) \in \mathcal{E}}(f_u - f_v)^2 \right)
    \label{eq:gmrf}
\end{equation}
The first term anchors each latent score at the prior mean $\mu_0$, while the second penalizes differences between adjacent scores, thereby inducing spatial correlation.
The parameters $\lambda_0 > 0$ and $\lambda_1 \geq 0$ control the strength of the prior anchoring and spatial coupling, respectively.
Since locations in $\mathcal{V} \setminus \mathcal{Z}$ are known to be traversable, we do not infer their states.
Instead, we fix their latent scores to a positive constant $c$, so they act as known traversable locations in the belief model.
Together, the resulting GMRF and the Bernoulli probit model induce agents' initial belief $p_0(\mathbf{X})$ over the unknown traversability states of locations.

Let $\mathcal{Z}_t \subseteq \mathcal{Z}$ denote the locations whose traversability remains unknown after observations at timestep $t$. When a location $z \in \mathcal{Z}_t$ is observed by an agent, it is removed from $\mathcal{Z}_t$ with its traversability belief $b_t(z)$ set to $1$ if traversable and $0$ otherwise.
To incorporate this observation while retaining Gaussian inference, we approximate the effect of the binary observation by fixing $f_z = +c$ when $z$ is traversable and $f_z = -c$ when it is blocked.
Through the pairwise terms in Eq.~\eqref{eq:gmrf}, these fixed values update beliefs at nearby unobserved locations.

Let $\mathcal{O}_t$ contain the uncertain locations observed at timestep $t$, together with their observed states.
We incorporate all observations in $\mathcal{O}_t$ into a single batch and use GaBP to estimate the latent marginals over $\mathcal{Z}_t$.
Let $\mu_v^t$ and $(\sigma_v^t)^2$ denote the resulting marginal mean and variance of $f_v$. 
For $v \in \mathcal{Z}_t$, integrating $\Phi(f_v)$ over the inferred Gaussian marginal gives the closed-form traversability estimate~\cite{rasmussen2006gaussian}
\begin{equation}
    b_t(v) \approx \Phi\!\left( \frac{\mu_v^t}{\sqrt{1 + (\sigma_v^t)^2}} \right)
    \label{eq:probit}
\end{equation}
Observed locations retain their known binary beliefs, while $b_t(v) = 1$ for $v \in \mathcal{V} \setminus \mathcal{Z}$.
These approximate estimates guide subsequent planning.
For a fixed marginal mean, a large marginal variance shifts the estimate toward $0.5$. Thus, poorly determined latent scores produce less confident traversability estimates.

After initial inference, we rerun GaBP only when new observations are obtained and reuse the previous estimates otherwise. 
We use message damping, which blends new and previous message parameters, to aid convergence~\cite{su2015convergence}. 
Each update stops when the largest message residual falls below a tolerance or after $K$ sweeps, using the final marginal estimates.
At convergence, GaBP gives exact means for the clamped Gaussian surrogate, while variances may be approximate on graphs with cycles~\cite{weiss2001correctness}.
Each sweep is linear in the number of variables and pairwise terms processed, with known locations contributing fixed evidence.


\subsection{Detour-Aware Traversal Costs}

The shared belief provides traversability estimates for individual locations.
We next convert these estimates into traversal costs for use by a MAPF planner. 
Let $\mathcal{G}_t = (\mathcal{V}_t, \mathcal{E}_t)$ denote the graph the planner searches at timestep $t$. 
It contains every location that has not yet been observed to be blocked. 
An observed blocked location is removed together with its incident edges, while locations in $\mathcal{Z}_t$ remain available.
Since unobserved locations are never removed, the realized traversable graph satisfies $\mathcal{G}^\star(\mathbf{x}) \subseteq \mathcal{G}_t$. 

We next map location beliefs to an edge-traversability estimate. 
For an edge $e = (u, v) \in \mathcal{E}_t$, we define
\begin{equation}
    \hat{b}_t(e) = \min\{\, b_t(u), b_t(v) \,\}
    \label{eq:edgebelief}
\end{equation}
The marginal traversability beliefs $b_t(u)$ and $b_t(v)$ do not determine the probability that both locations are traversable. 
Their product assumes independence, while computing the pairwise probability requires the joint posterior of the two latent variables.
Instead, we use the smaller marginal as a lightweight edge-traversability estimate that only requires the location beliefs already produced by GaBP. 
For each edge $e = (u, v) \in \mathcal{E}_t$ incident to a location in $\mathcal{Z}_t$, we compute
\begin{equation}
    C_{\mathrm{det}}^t(e) = d_{\mathcal{G}_t \setminus \{e\}}(u, v)
    \label{eq:cdet}
\end{equation}
where $d$ denotes the unweighted single-agent shortest-path distance.
Thus, $C_{\mathrm{det}}^t(e)$ is the length of the shortest edge-bypass distance from $u$ to $v$.
This quantity depends only on the current graph and the edge, and is shared across the agents.
If removing $e$ disconnects $u$ from $v$, we set $C_{\mathrm{det}}^t(e) = |\mathcal{V}|$.
Since any finite unweighted shortest path in $\mathcal{G}_t$ does not need to revisit a location, its length is at most $|\mathcal{V}| - 1$.
Thus, this value exceeds every finite detour length and provides a finite penalty when no detour exists.

We combine the estimated traversability with the detour distance to define the detour-aware edge cost for each edge $e \in \mathcal{E}_t$ incident to a location in $\mathcal{Z}_t$
\begin{equation}
    c_t(e) = \hat{b}_t(e) \cdot 1 + \bigl(1 - \hat{b}_t(e) \bigr)\, C_{\mathrm{det}}^t(e)
    \label{eq:softcost}
\end{equation}
This cost interpolates between the unit traversal cost and the local detour distance. 
All remaining edges in $\mathcal{E}_t$ have $\hat{b}_t(e) = 1$ and therefore retain unit cost, so no detour needs to be computed for them. 
Wait actions also retain unit cost. 
An edge incident to an unobserved, uncertain location with a short alternative remains inexpensive, while an edge with a long detour between its endpoints receives a higher cost. 
Since both terms are measured in steps, the construction introduces no additional weighting parameter.
Because $\hat{b}_t(e)$ is an edge-level estimate and $C_{\mathrm{det}}^t(e)$ considers only removal of $e$, $c_t(e)$ is a local detour-aware surrogate rather than the exact expected cost of future execution.

\begin{algorithm}[t]
\caption{\methodname execution loop}
\label{alg:bamapf}
\begin{algorithmic}[1]
    \parState{Initialize the planning graph from $\mathcal{G}$, and the belief and costs from the prior $p_0$}
    \State $\mathbf{q}^0 \gets (s_1,\dots,s_n), \; \mathcal{J} \gets \varnothing, \; t \gets 0$ \Comment{current joint plan}
    \While{$\mathbf{q}^t \neq \mathbf{g}$}
        \parState{$\mathcal{O}_t \gets$ newly observed adjacent uncertain locations and their shared states}
        \If{$\mathcal{O}_t \neq \varnothing$}
            \parState{Incorporate $\mathcal{O}_t$ and update the posterior with GaBP}
            \State Update $\mathcal{G}_t$, detours, and $c_t$
            \State $\mathcal{J} \gets \varnothing$ \Comment{planner input changed}
        \EndIf
         \If{$\mathcal{J}$ has no valid next action}
            \State $\mathcal{J} \gets \textsc{MAPF}(\mathcal{G}_t,\, c_t,\, \mathbf{q}^t,\, \mathbf{g})$
            \If{$\mathcal{J}$ has no valid next action}
                \State \Return Failure
            \EndIf
        \EndIf
        \State Execute the first joint action of $\mathcal{J}$
        \State Update $\mathbf{q}^{t+1}$
        \State Remove the executed joint action from $\mathcal{J}$
        \State $t \gets t + 1$
    \EndWhile
\end{algorithmic}
\end{algorithm}



We cache each detour length along with the corresponding shortest detour path, when one exists. 
The detour lengths depend only on $\mathcal{G}_t$ and not on the posterior beliefs, so a traversable observation leaves them unchanged.
When an uncertain location is observed to be blocked, its removal invalidates only the cached detour paths that use one of the newly removed edges. 
We recompute invalidated detours and reuse surviving cached detour paths, which remain shortest because $\mathcal{G}_t$ only loses edges and vertices.
Without such reuse, computing all detours requires one unweighted shortest-path query for each edge incident to a location in $\mathcal{Z}_t$.

\subsection{Planning and Execution}

The resulting weighted graph defines the deterministic MAPF instance solved at each replanning step. 
At each step, the MAPF planner receives the current configuration $\mathbf{q}^t = (q_1^t, \dots, q_n^t)$, the goals $\mathbf{g} = (g_1, \dots, g_n)$, the graph $\mathcal{G}_t$, and the edge costs $c_t$. 
Search-based MAPF planners such as CBS~\cite{sharon2015cbs} use $c_t$ as edge costs to find paths for agents. 
MAPF planners such as PIBT~\cite{okumura2022pibt} and LaCAM~\cite{okumura2023lacam} use $c_t$ to estimate the weighted distance-to-go for selecting candidate moves. 
In general, the same belief and cost model can be applied to different MAPF planners.


Algorithm~\ref{alg:bamapf} summarizes the execution loop.
Let $\mathcal{J}$ denote the current joint plan, represented as an ordered sequence of joint actions returned by the underlying MAPF planner. 
Replanning is triggered by any new observation, not only by a blocked one, because even a traversable observation can update posterior beliefs and, in turn, the edge costs.
A MAPF planner that returns a single joint action rather than a full joint plan is queried again at the next timestep. Execution terminates with failure when a time limit is reached.

A valid joint action consists of waits or adjacent moves that avoid known blocked locations and vertex and edge conflicts.
Because adjacent uncertain locations are observed before moving, noiseless observations reveal any blocked destination before entry. 
With a sound planner, executed actions are therefore valid in $\mathcal{G}^\star(\mathbf{x})$. 

\section{EXPERIMENTAL EVALUATION}
\label{sec:experiments}

We organize our evaluations around three questions.
\begin{enumerate}
    \item[\textbf{Q1}] Does \methodname improve execution under uncertain traversability?
    \item[\textbf{Q2}] Does the benefit persist across planner families and problem scales?
    \item[\textbf{Q3}] How sensitive is \methodname to the spatial information encoded by the prior $p_0(\mathbf{X})$?
\end{enumerate}

\subsection{Experimental Setup}
\label{sec:exp_setup}

\paragraph{Benchmark maps}
We evaluate on the standard MAPF benchmark~\cite{stern2019mapf} using maps spanning open, random, maze, room, warehouse, game, and city layouts, grouped into four scale tiers, as summarized in Table~\ref{tab:benchmark_maps}.

\begin{table}[t]
    \vspace*{5pt}
    \centering
    \caption{Benchmark maps and team sizes $n$ used for evaluation.}
    \label{tab:benchmark_maps}
    \small
    \setlength{\tabcolsep}{4pt}
    \begin{tabular}{@{}llr@{}}
        \toprule
        Tier & Maps & $n$\\
        \midrule
        Small &
        \begin{tabular}[t]{@{}l@{}}
            \texttt{empty-32-32}, \texttt{random-32-32-10}, \\
            \texttt{maze-32-32-4}, \texttt{random-32-32-20}, \\
            \texttt{room-32-32-4}
        \end{tabular}
        & 10 \\
        \addlinespace[3pt]
        Medium &
        \begin{tabular}[t]{@{}l@{}}
            \texttt{den312d}, \texttt{empty-48-48}, \\
            \texttt{room-64-64-8}, \texttt{random-64-64-20}
        \end{tabular}
        & 50 \\
        \addlinespace[3pt]
        Large &
        \begin{tabular}[t]{@{}l@{}}
            \texttt{warehouse-10-20-10-2-2}, \\
            \texttt{den520d}, \texttt{maze-128-128-10}, \\
            \texttt{lt\_gallowstemplar\_n}
        \end{tabular}
        & 200 \\
        \addlinespace[3pt]
        Huge &
        \begin{tabular}[t]{@{}l@{}}
            \texttt{warehouse-20-40-10-2-2}, \\
            \texttt{brc202d}, \texttt{Berlin\_1\_256}, \\
            \texttt{orz900d}
        \end{tabular}
        & 800 \\
        \bottomrule
    \end{tabular}
\end{table}

\begin{figure}
    \centering
    \includegraphics[width=0.85\linewidth]{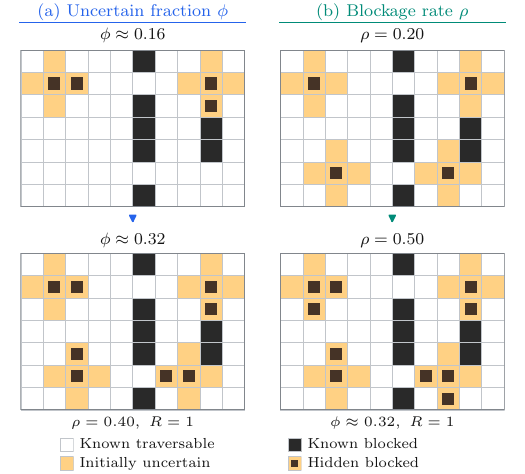}
    \caption{\textbf{Illustration of the uncertainty parameters used in the experiments.} 
    Orange locations have initially uncertain traversability, and inset dark squares indicate locations that are blocked in the hidden realization.
    (a) Increasing $\phi$ increases the portion of the map whose traversability is initially unknown.
    (b) A higher $\rho$ corresponds to a larger fraction of uncertain locations being blocked.}
    \label{fig:uncertainty_parameters}
\end{figure}

\paragraph{Uncertain environment generation}
We instantiate the environment distribution $\mathcal{P}$ using spatially structured hidden obstacles. 
We first sample uncertainty centers from the traversable locations of each benchmark map. 
For environment generation, let $d(v)$ denote the unweighted distance in $\mathcal{G}$ from location $v$ to its nearest uncertainty center.
For an uncertainty radius $R$, locations other than starts and goals satisfying $d(v) \leq R$ form the uncertain set $\mathcal{Z}$. 
Before feasibility filtering, the hidden state of each $v \in \mathcal{Z}$ is sampled independently, conditioned on the sampled centers, with
\begin{equation}
    \Pr[X_v = 0 \mid \mathrm{sampled\, centers}] = \exp\left(-\frac{d(v)^2}{2\ell^2}\right)
    \label{eq:hazard_kernel}
\end{equation}
where $\ell > 0$ controls how quickly blockage probability decreases with distance from an uncertainty center. 
We characterize each realization by the uncertain fraction $\phi = \frac{|\mathcal{Z}|}{|\mathcal{V}|}$ and blockage rate $\rho = \frac{|\{v \in \mathcal{Z}: x_v = 0\}|}{|\mathcal{Z}|}$. 
We choose the number of uncertainty centers and $\ell$ to target specified values of $\phi$ and $\rho$.
For settings where $\rho = 0$, we retain $\mathcal{Z}$ but set every location in it to be traversable.
Additionally, when $\phi = 0$, there are no uncertain locations and $\rho$ is not used. 
Fig.~\ref{fig:uncertainty_parameters} illustrates the effect of these parameters.
We retain only realizations for which $\mathcal{G}^{\star}(\mathbf{x})$ admits a collision-free solution, and the accepted realization $\mathbf{x}$ remains fixed throughout execution.
In the standard benchmark runs, agents know $\mathcal{Z}$ but not the uncertainty centers, the blockage probabilities in Eq.~\eqref{eq:hazard_kernel}, or the realization $\mathbf{x}$.

\begin{figure*}[t]
    \centering
    \includegraphics[width=0.85\textwidth]{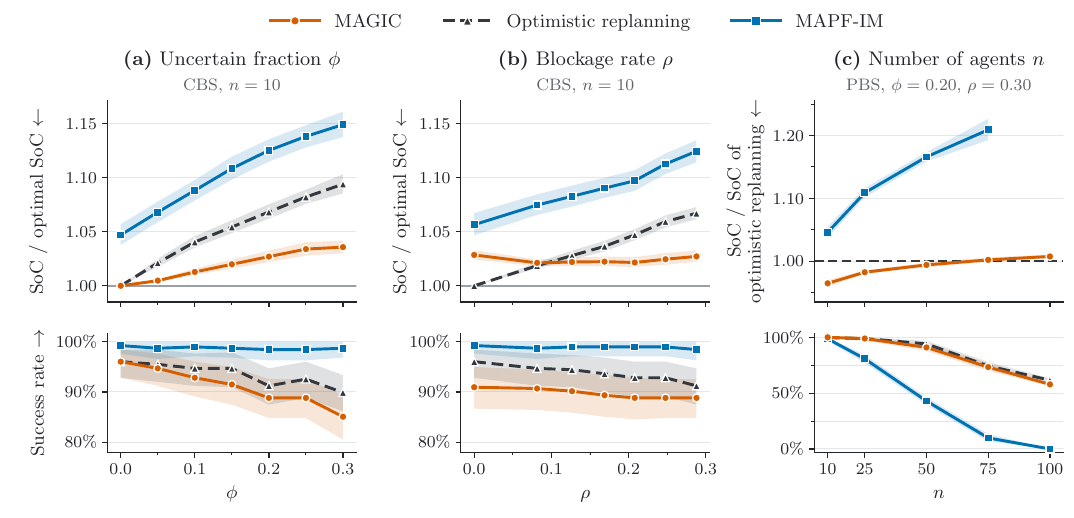}
    \caption{\textbf{Executed SoC and success rate under uncertain traversability.} 
    \methodname and optimistic replanning use CBS with $n = 10$ in panels (a) and (b), and PBS in (c).
    Cost ratios use optimal full-information SoC obtained by CBS on $\mathcal{G}^{\star}(\mathbf{x})$ as the reference in (a,b), and optimistic replanning in (c). 
    Panel (a) varies $\phi$ and (b) varies $\rho$.
    Each cost curve uses instances completed by the method and its reference, whereas success rates include all attempts.
    Shading denotes 95\% confidence intervals. 
    MAPF-IM has no successful runs at $n = 100$. 
    Lower executed SoC and higher success rate are better.}
    \label{fig:main_benchmark}
\end{figure*}

\paragraph{Experimental conditions}
Unless varied, the generator targets an uncertain fraction $\phi = 0.20$, a blockage rate $\rho = 0.30$, and an uncertainty radius $R = 3$. 
To answer \textbf{Q1}, the small-tier experiments vary the target values of $\phi$ and $\rho$ from $0.00$ to $0.30$ in increments of $0.05$. 
The small-tier team size study varies $n \in \{10, 25, 50, 75, 100\}$ while holding the uncertainty parameters fixed. 
Each small-tier setting uses 25 benchmark scenarios per map and three independently sampled realizations $\mathbf{x}$ per scenario, yielding 375 instances across the five maps.
Each medium-, large-, and huge-tier setting uses 25 scenarios per map and one realization per scenario, yielding 100 instances.
To answer \textbf{Q2}, we evaluate $n = 10, 50, 200$, and $800$ agents on the small-, medium-, large-, and huge-tiers, respectively. 

\paragraph{Planners and comparisons}
We compare \methodname with location-adapted MAPF-IM and with optimistic replanning. 
The MAPF-IM comparison evaluates \methodname against the closest prior approach to our setting.
Optimistic replanning uses the same realization $\mathbf{x}$, start-goal assignment, and underlying MAPF planner as \methodname, but treats every unresolved location $ v \in \mathcal{Z}_t$ as traversable with unit edge costs. 
This comparison evaluates belief-guided planning together with its observation-triggered replanning policy.
For planners that maintain a joint plan, optimistic replanning replans when a newly observed location is blocked, whereas \methodname replans after every new observation. 
PIBT, which returns a single joint action, is queried at each timestep. 

The small-tier experiments use CBS, EECBS with suboptimality bounds $w = 1.05$ and $w = 1.10$~\cite{li2021eecbs}, and PBS~\cite{ma2019pbs}. 
The medium- and large-tiers use PBS, PIBT, PIBT+~\cite{okumura2022pibt}, and LaCAM. 
The huge-tier uses PIBT, PIBT+, and LaCAM.
For each small-tier instance, we also run CBS with full 
knowledge of $\mathcal{G}^{\star}(\mathbf{x})$. When the full-information CBS run succeeds,
its optimal SoC provides a common lower bound for the planners evaluated on that instance. 
For the medium-, large-, and huge-tiers, our primary cost comparison is instead between \methodname and optimistic replanning using the same underlying planner. 
We additionally compare against MAPF-IM adapted to location-based uncertainty~\cite{malka2024mapfim}. 
During local conflict resolution, its search adds a fixed penalty to the distance estimate for transitions incident to $\mathcal{Z}_t$, rather than using $b_t(v)$ maintained by \methodname. 

\paragraph{Inference and limits}
Unless otherwise stated, we set $\mu_0 = 0$, $\lambda_0 = \lambda_1 = 1$, and $c = 2$ for \methodname. For GaBP, we use a message damping factor of $0.5$, a residual tolerance of $10^{-8}$, and a maximum of $K = 200$ sweeps per belief update.
The small-, medium-, large-, and huge-tiers use per-call planner time limits of $60, 120, 120, 300$ s, total runtime limits of $300, 3600, 3600, 14400$ s, and execution limits of $500, 3000, 3000, 12000$ timesteps, respectively.
Methods compared within the same experimental condition are subject to the same limits.
All experiments were run on a workstation with an Intel Core i9-14900K CPU, 32 logical CPUs, and 62 GiB of usable memory.
Independent runs were executed in parallel.

\paragraph{Evaluation metrics}
We report the executed SoC and the success rate separately.
Cost ratios are computed per instance and averaged over instances completed by both the method and its stated reference.
Each figure or table specifies that reference.
Success rates include all attempted instances, counting runs that fail to complete within the stated limits or violate traversability or collision constraints as failures.
We additionally report 95\% confidence intervals.

\begin{table*}[t]
    \vspace*{5pt}
    \centering
    \caption{\textbf{Performance across planner families and problem scales at
    $\mathbf{\phi = 0.20}$, $\mathbf{\rho = 0.30}$, and $\mathbf{R = 3}$.}}
    \label{tab:cross-scale}

    \small
    \setlength{\tabcolsep}{3.5pt}

    \settowidth{\SuccessHalf}{Success $\uparrow$}%
    \settowidth{\dimen0}{/}%
    \addtolength{\SuccessHalf}{-\dimen0}%
    \setlength{\SuccessHalf}{0.5\SuccessHalf}%

    \begin{tabular}{
        @{}l@{\hspace{1.1em}}%
        c r@{/}l@{\hspace{1.1em}}%
        c w{r}{\SuccessHalf}@{/}w{l}{\SuccessHalf}@{\hspace{1.1em}}%
        c w{r}{\SuccessHalf}@{/}w{l}{\SuccessHalf}@{\hspace{1.1em}}%
        c w{r}{\SuccessHalf}@{/}w{l}{\SuccessHalf}@{}
    }
        \toprule

        & \multicolumn{3}{c@{\hspace{1.1em}}}{Small ($n=10$)}
        & \multicolumn{3}{c@{\hspace{1.1em}}}{Medium ($n=50$)}
        & \multicolumn{3}{c@{\hspace{1.1em}}}{Large ($n=200$)}
        & \multicolumn{3}{c@{}}{Huge ($n=800$)}
        \\

        \cmidrule(lr){2-4}
        \cmidrule(lr){5-7}
        \cmidrule(lr){8-10}
        \cmidrule(lr){11-13}

        Planner
        & SoC ratio $\downarrow$
        & \multicolumn{2}{c@{\hspace{1.1em}}}{Success $\uparrow$}
        & SoC ratio $\downarrow$
        & \multicolumn{2}{c@{\hspace{1.1em}}}{Success $\uparrow$}
        & SoC ratio $\downarrow$
        & \multicolumn{2}{c@{\hspace{1.1em}}}{Success $\uparrow$}
        & SoC ratio $\downarrow$
        & \multicolumn{2}{c@{}}{Success $\uparrow$}
        \\

        \midrule

        CBS
        & $0.966 \pm 0.004$ & 91.2 & 88.8
        & \multicolumn{3}{c@{\hspace{1.1em}}}{--}
        & \multicolumn{3}{c@{\hspace{1.1em}}}{--}
        & \multicolumn{3}{c@{}}{--}
        \\

        EECBS ($w=1.05$)
        & $0.967 \pm 0.005$ & 96.8 & 94.9
        & \multicolumn{3}{c@{\hspace{1.1em}}}{--}
        & \multicolumn{3}{c@{\hspace{1.1em}}}{--}
        & \multicolumn{3}{c@{}}{--}
        \\

        EECBS ($w=1.10$)
        & $0.969 \pm 0.005$ & 97.6 & 96.5
        & \multicolumn{3}{c@{\hspace{1.1em}}}{--}
        & \multicolumn{3}{c@{\hspace{1.1em}}}{--}
        & \multicolumn{3}{c@{}}{--}
        \\

        PBS
        & $0.963 \pm 0.006$ & 100.0 & 99.7
        & $0.987 \pm 0.004$ & 100   & 99
        & $0.990 \pm 0.002$ & 95    & 74
        & \multicolumn{3}{c@{}}{--}
        \\

        LaCAM
        & \multicolumn{3}{c@{\hspace{1.1em}}}{--}
        & $0.986 \pm 0.004$ & 100 & 100
        & $0.989 \pm 0.001$ & 100 & 100
        & $0.995 \pm 0.001$ & 100 & 94
        \\

        PIBT+
        & \multicolumn{3}{c@{\hspace{1.1em}}}{--}
        & $0.995 \pm 0.009$ & 100 & 100
        & $0.991 \pm 0.003$ & 100 & 100
        & $0.995 \pm 0.001$ & 100 & 100
        \\

        PIBT
        & \multicolumn{3}{c@{\hspace{1.1em}}}{--}
        & $0.991 \pm 0.008$ & 96 & 96
        & $0.992 \pm 0.003$ & 95 & 94
        & $0.993 \pm 0.002$ & 79 & 55
        \\

        \midrule

        MAPF-IM$^{\dagger}$
        & $1.052 \pm 0.009$ & 91.2 & 98.4
        & $1.103 \pm 0.010$ & 100  & 66
        & $1.209 \pm 0.023$ & 100  & 11
        & \multicolumn{3}{c@{}}{--}
        \\

        \bottomrule
    \end{tabular}
\end{table*}

\subsection{Results and Discussion}

\paragraph{\textbf{Q1}}

Fig.~\ref{fig:main_benchmark} shows how executed SoC and success rate vary with $\phi$, $\rho$, and $n$.
As the uncertain fraction $\phi$ increases, optimistic replanning moves farther from the full-information optimum.
At $\phi = 0$, \methodname and optimistic replanning both recover the optimal full-information SoC because $\mathcal{Z} = \varnothing$.
At $\phi = 0.30$, optimistic replanning is $9.4\%$ above the optimum, whereas \methodname is $3.6\%$ above it.
MAPF-IM is $14.9\%$ above the optimum at the same setting.
Thus, as more locations are uncertain to traverse, \methodname recovers a substantial portion of the execution-cost gap between optimistic replanning and the full-information optimum.

The blockage rate sweep shows when the additional caution introduced by \methodname becomes useful.
At $\rho = 0$, every location in $\mathcal{Z}$ is traversable in the realized environment, so optimistic replanning matches the full-information optimum.
\methodname is $2.9\%$ above the optimum because it still penalizes transitions involving unresolved locations.
As the blockage rate increases, this cost is offset by avoiding routes through locations that are more likely to be blocked.
At the largest tested $\rho$, \methodname is $2.7\%$ above the optimum, compared with $6.7\%$ for optimistic replanning and $12.4\%$ for MAPF-IM.


The executed SoC advantage of \methodname decreases as \(n\) increases. 
This trend is consistent with two effects. 
Larger teams can gather observations in parallel, shortening the period during which inference about $\mathcal Z_t$ can influence route selection, while increased coordination can also leave fewer alternative routes around uncertain locations. 
Accordingly, \methodname and optimistic replanning approach parity as $n$ grows, while MAPF-IM exhibits a sharper decline in success and has no successful run at $n = 100$.

\paragraph{\textbf{Q2}}

\begin{figure}
    \centering
    \includegraphics[width=0.85\linewidth]{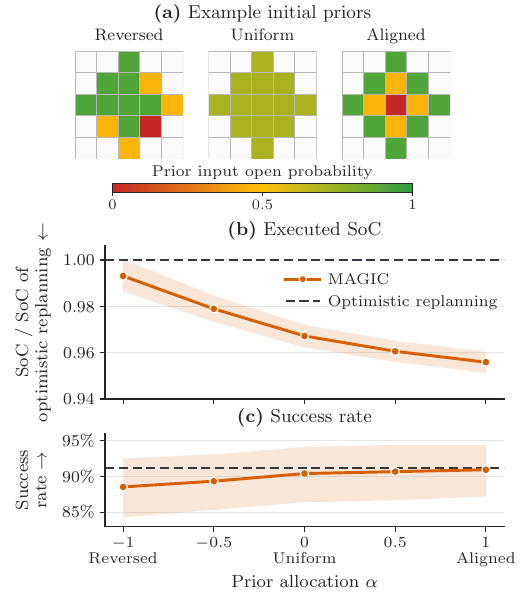}
    \caption{\textbf{Sensitivity to the spatial allocation of the initial prior}
    (a) Illustration of reversed, uniform, and aligned prior traversability probabilities.
    A higher prior blockage probability corresponds to a lower initial traversability belief.
    (b) Executed SoC relative to optimistic replanning as the prior allocation varies from reversed to aligned.
    (c) Success rate over all attempted instances.
    Shading denotes 95\% confidence intervals. 
    Lower executed SoC and higher success rates are better.
    }
    \label{fig:prior_sensitivity}
\end{figure}

Table~\ref{tab:cross-scale} evaluates \textbf{Q2} across the planner families and four scale tiers.
For each planner and tier, the table reports the mean per-instance executed-SoC ratio and reference/method success rates over all attempts.
Ratios below one favor the evaluated method and above one favor the reference.
For each non-daggered row, the SoC ratio compares \methodname with optimistic replanning using the same underlying planner.
The $\pm$ value reports the wider side of the 95\% bootstrap intervals.
For MAPF-IM$^{\dagger}$, the reference is optimistic replanning with CBS for the small tier and with LaCAM for the other tiers.

Across the small-, medium-, and large-tiers, \methodname achieves lower executed SoC than MAPF-IM on $96.3\%$. 
All 15 \methodname configurations have mean ratios below one.
The smaller gains in the larger tiers are consistent with the trend discussed in \textbf{Q1}, but maps, team size, and resource limits also differ across tiers. 
At $n = 800$, PIBT+ retains $100\%$ success under both methods with a mean SoC ratio of $0.995$.
The success rate results show that this cost benefit does not imply uniform reliability among planners.
\methodname solves the same underlying instances but performs additional GaBP inference and belief-dependent planning, while planners that return complete plans may also fully replan after traversable observations. 
This more demanding online workload can cause difficult runs to exhaust planner or total-runtime limits before all agents reach their goals. 
Thus, the lower success rates in some settings primarily reflect a computational tradeoff.
MAPF-IM exhibits a different tradeoff.
Its higher success rate on the small-tier is consistent with its impact-detection and localized conflict resolution, but its success rate decreases substantially at larger scales.

We additionally varied the uncertainty radius $R$ and found little aggregate sensitivity to it.
On the large warehouse map, however, \methodname incurred a higher executed SoC than optimistic replanning at $R = 1$, but for $R \geq 2$ the trend reversed.
This suggests that very localized uncertainty may provide too little spatial evidence to offset conservative detours in narrow aisles, whereas larger regions of uncertainty make neighboring observations more informative. 
Further, medium- and large-tier sweeps over $\phi$ and $\rho$ broadly reproduced similar trends. 

\paragraph{\textbf{Q3}}

\textbf{Q1} and \textbf{Q2} use the same prior model and parameter settings. 
To answer \textbf{Q3}, we now vary which locations in $\mathcal{Z}$ receive higher or lower prior traversability probabilities.
Using the same small-tier instances at $\phi = 0.20$, $\rho = 0.30$ and $R = 3$, we construct five initial priors indexed by $\alpha \in \{-1, -0.5, 0, 0.5, 1\}$. 
The aligned prior $\alpha = 1$ assigns each location a prior traversability probability consistent with its probability under the environment generator.
The reversed prior $\alpha=-1$ uses the same probability values but assigns them in reverse order, so locations that are more likely to be traversable under the generator receive lower prior traversability probabilities, and vice versa.
At $\alpha = 0$, every location in $\mathcal{Z}$ receives the same average traversability probability.
The intermediate settings $\alpha = \pm0.5$ shift the corresponding aligned or reversed probabilities by half toward this average.
To obtain these probabilities, we replace the shared mean $\mu_0$ in Eq.~\eqref{eq:gmrf} with location-specific prior means chosen to produce the corresponding initial traversability probabilities, while keeping the other settings unchanged. 
Fig.~\ref{fig:prior_sensitivity}(a) illustrates these effects.

Fig.~\ref{fig:prior_sensitivity}(b) shows a consistent reduction in executed SoC as the spatial allocation becomes better aligned with the generating probability profile. 
When comparing the aligned and reversed priors directly, the aligned prior reduces executed SoC by $3.40\%$ for successful instances. 
The success rate results show a similar, though small, trend.
It increases from $88.5\%$ under the reversed prior to $90.9\%$ under the aligned prior, compared with $91.2\%$ for the optimistic replanning on the same benchmark set.
The aligned prior does not improve every individual instance.
Still, its aggregate advantage is strongest on maps with constrained routes such as rooms, where assigning a low prior traversability probability to useful passages can make them artificially expensive.
The result, therefore, shows that informative spatial structure in $p_0(\mathbf{X})$ can improve execution. 

\paragraph{Physical robot demonstration}
We also demonstrate \methodname on a team of two TurtleBots in a fixed indoor environment. 
Planning observations are restricted to adjacent locations to match the observation model.
In the demonstration, one robot observes a blocked, uncertain location, which updates the shared belief at a nearby location that remains unobserved by either robot.
This causes the second robot to change its route before directly observing that location. 
A supplemental video shows the complete executions together with additional qualitative examples.

\section{CONCLUSION}

We formulated \problemname for fixed environments with initially uncertain traversability and introduced \methodname, which updates a shared spatial belief from local observations and converts inferred traversability into detour-aware costs for standard MAPF planners.
Across different tiers, \methodname achieves lower executed SoC than existing approaches on $96.3\%$ of instances.
The benefit persists across several planner families and scales to teams of $800$ agents. 
We further find that execution depends on how the prior $p_0(\mathbf{X})$ assigns traversability probabilities across $\mathcal{Z}$. 
Future work will extend \problemname to lifelong tasks and environments whose traversability changes during execution.

\bibliographystyle{IEEEtran}
\bibliography{IEEEabrv,references}

@inproceedings{stern2019mapf,
    author = {Stern, Roni and Sturtevant, Nathan R. and Felner, Ariel and Koenig, Sven and Ma, Hang and
              Walker, Thayne T. and Li, Jiaoyang and Atzmon, Dor and Cohen, Liron and
              Kumar, T. K. Satish and Bart{\'a}k, Roman and Boyarski, Eli},
    title = {Multi-Agent Pathfinding: Definitions, Variants, and Benchmarks},
    booktitle = {Proc. Symp. Combin. Search (SoCS)},
    pages = {151--158},
    year = {2019},
    doi = {10.1609/socs.v10i1.18510},
}

@article{wurman2008coordinating,
    author = {Wurman, Peter R. and D'Andrea, Raffaello and Mountz, Mick},
    title = {Coordinating Hundreds of Cooperative, Autonomous Vehicles in Warehouses},
    journal = {AI Mag.},
    volume = {29},
    number = {1},
    pages = {9--19},
    year = {2008},
    doi = {10.1609/aimag.v29i1.2082},
}

@inproceedings{shofer2023mapfou,
    author = {Shofer, Bar and Shani, Guy and Stern, Roni},
    title = {Multi Agent Path Finding under Obstacle Uncertainty},
    booktitle = {Proc. Int. Conf. Autom. Plan. Sched. (ICAPS)},
    volume = {33},
    pages = {402--410},
    year = {2023},
    doi = {10.1609/icaps.v33i1.27219},
}

@inproceedings{malka2024mapfim,
    author = {Malka, Nir and Shani, Guy and Stern, Roni},
    title = {Online Planning for Multi Agent Path Finding in Inaccurate Maps},
    booktitle = {Proc. IEEE/RSJ Int. Conf. Intell. Robots Syst. (IROS)},
    pages = {10214--10221},
    year = {2024},
    doi = {10.1109/IROS58592.2024.10801975},
}

@article{ocallaghan2012gpom,
    author = {O'Callaghan, Simon T. and Ramos, Fabio T.},
    title = {Gaussian Process Occupancy Maps},
    journal = {Int. J. Robot. Res.},
    volume = {31},
    number = {1},
    pages = {42--62},
    year = {2012},
    doi = {10.1177/0278364911421039},
}

@article{weiss2001correctness,
    author = {Weiss, Yair and Freeman, William T.},
    title = {Correctness of Belief Propagation in {Gaussian} Graphical Models of Arbitrary Topology},
    journal = {Neural Comput.},
    volume = {13},
    number = {10},
    pages = {2173--2200},
    year = {2001},
    doi = {10.1162/089976601750541769},
}

@inproceedings{zhang2024ggo,
    author = {Zhang, Yulun and Jiang, He and Bhatt, Varun and Nikolaidis, Stefanos and Li, Jiaoyang},
    title = {Guidance Graph Optimization for Lifelong Multi-Agent Path Finding},
    booktitle = {Proc. Int. Joint Conf. Artif. Intell. (IJCAI)},
    pages = {311--320},
    year = {2024},
    doi = {10.24963/ijcai.2024/35},
}

@inproceedings{li2021rhcr,
    author = {Li, Jiaoyang and Tinka, Andrew and Kiesel, Scott and Durham, Joseph W. and
              Kumar, T. K. Satish and Koenig, Sven},
    title = {Lifelong Multi-Agent Path Finding in Large-Scale Warehouses},
    booktitle = {Proc. AAAI Conf. Artif. Intell.},
    volume = {35},
    number = {13},
    pages = {11272--11281},
    year = {2021},
    doi = {10.1609/aaai.v35i13.17344},
}

@inproceedings{nikolova2008ctp,
    author = {Nikolova, Evdokia and Karger, David R.},
    title = {Route Planning under Uncertainty: The {Canadian Traveller Problem}},
    booktitle = {Proc. AAAI Conf. Artif. Intell.},
    pages = {969--974},
    year = {2008},
}

@inproceedings{stadler2023collaborative,
    author = {Stadler, Martina and Banfi, Jacopo and Roy, Nicholas},
    title = {Approximating the Value of Collaborative Team Actions for Efficient Multiagent Navigation in Uncertain Graphs},
    booktitle = {Proc. Int. Conf. Autom. Plan. Sched. (ICAPS)},
    volume = {33},
    number = {1},
    pages = {677--685},
    year = {2023},
    doi = {10.1609/icaps.v33i1.27250},
}

@inproceedings{veys2024sparse,
    author = {Veys, Yasmin and Kurtz, Martina Stadler and Roy, Nicholas},
    title = {Generating Sparse Probabilistic Graphs for Efficient Planning in Uncertain Environments},
    booktitle = {Proc. IEEE Int. Conf. Robot. Autom. (ICRA)},
    pages = {133--139},
    year = {2024},
    doi = {10.1109/ICRA57147.2024.10610493},
}

@inproceedings{silver2010pomcp,
    author = {Silver, David and Veness, Joel},
    title = {{Monte-Carlo} Planning in Large {POMDPs}},
    booktitle = {Adv. Neural Inf. Process. Syst.},
    volume = {23},
    pages = {2164--2172},
    year = {2010},
}

@inproceedings{regev2016belief,
    author = {Regev, Tal and Indelman, Vadim},
    title = {Multi-Robot Decentralized Belief Space Planning in Unknown Environments via Efficient Re-Evaluation of Impacted Paths},
    booktitle = {Proc. IEEE/RSJ Int. Conf. Intell. Robots Syst. (IROS)},
    pages = {5591--5598},
    year = {2016},
    doi = {10.1109/IROS.2016.7759822},
}

@inproceedings{shankar2020mrfmap,
    author = {Shankar, Kumar Shaurya and Michael, Nathan},
    title = {{MRFMap}: Online Probabilistic {3D} Mapping Using Forward Ray Sensor Models},
    booktitle = {Proc. Robotics: Science and Systems (RSS)},
    year = {2020},
    doi = {10.15607/RSS.2020.XVI.060},
}

@misc{zhou2025scos,
    author = {Zhou, Li and Ceyhan, Elvan},
    title = {Stochastic Path Planning in Correlated Obstacle Fields},
    year = {2025},
    note = {arXiv:2509.19559},
}

@article{sharon2015cbs,
    author = {Guni Sharon and Roni Stern and Ariel Felner and Nathan R. Sturtevant},
    title = {Conflict-based search for optimal multi-agent pathfinding},
    journal = {Artif. Intell.},
    volume = {219},
    pages = {40--66},
    year = {2015},
    doi = {10.1016/j.artint.2014.11.006},
}

@article{okumura2022pibt,
    author = {Okumura, Keisuke and Machida, Manao and D{\'e}fago, Xavier and Tamura, Yasumasa},
    title = {Priority inheritance with backtracking for iterative multi-agent path finding},
    journal = {Artif. Intell.},
    volume = {310},
    pages = {103752},
    year = {2022},
    doi = {10.1016/j.artint.2022.103752},
}

@inproceedings{okumura2023lacam,
    author = {Okumura, Keisuke},
    title = {{LaCAM}: Search-Based Algorithm for Quick Multi-Agent Pathfinding},
    booktitle = {Proc. AAAI Conf. Artif. Intell.},
    volume = {37},
    number = {10},
    pages = {11655--11662},
    year = {2023},
    doi = {10.1609/aaai.v37i10.26377},
}

@inproceedings{li2021eecbs,
    author = {Li, Jiaoyang and Ruml, Wheeler and Koenig, Sven},
    title = {{EECBS}: A Bounded-Suboptimal Search for Multi-Agent Path Finding},
    booktitle = {Proc. AAAI Conf. Artif. Intell.},
    volume = {35},
    number = {14},
    pages = {12353--12362},
    year = {2021},
    doi = {10.1609/aaai.v35i14.17466},
}

@inproceedings{ma2019pbs,
    author = {Ma, Hang and Harabor, Daniel and Stuckey, Peter J. and Li, Jiaoyang and Koenig, Sven},
    title = {Searching with Consistent Prioritization for Multi-Agent Path Finding},
    booktitle = {Proc. AAAI Conf. Artif. Intell.},
    volume = {33},
    number = {1},
    pages = {7643--7650},
    year = {2019},
    doi = {10.1609/aaai.v33i01.33017643},
}

@book{rasmussen2006gaussian,
    author = {Rasmussen, Carl Edward and Williams, Christopher K. I.},
    title = {Gaussian Processes for Machine Learning},
    publisher = {The MIT Press},
    year = {2006},
}

@article{su2015convergence,
    author = {Su, Qinliang and Wu, Yik-Chung},
    title = {On Convergence Conditions of {Gaussian} Belief Propagation},
    journal = {IEEE Trans. Signal Process.},
    volume = {63},
    number = {5},
    pages = {1144--1155},
    year = {2015},
    doi = {10.1109/TSP.2015.2389755},
}

\end{document}